\documentclass[lettersize,journal]{IEEEtran}

\usepackage{amssymb}
\usepackage{array}
\usepackage{makecell}

\usepackage{pifont}
\usepackage{xcolor, colortbl}
\usepackage{amsmath,amsfonts,amssymb,amsthm}
\usepackage{mathtools}
\usepackage{commath}
\usepackage{multirow}

\usepackage{xcolor}
\usepackage{graphbox}
\usepackage{overpic}
\usepackage[pagebackref=true,breaklinks=true,colorlinks,bookmarks=false]{hyperref}

\def\eg{\emph{e.g.}}
\def\ie{\emph{i.e.}}
\newcommand{\xmark}{\ding{55}}%

\begin{document}

\title{Monitoring social distancing \\ with single image depth estimation}

\author{Alessio~Mingozzi,
        Andrea~Conti,
        Filippo~Aleotti, \\
        Matteo~Poggi, 
        Stefano~Mattoccia$^*$
\thanks{A. Mingozzi, A. Conti, F. Aleotti, M. Poggi and S. Mattoccia are with the Department of Computer Science and Engineering (DISI), University of Bologna, 40136, Italy.} 
\thanks{$^*$E-mail: stefano.mattoccia@unibo.it}
}

\maketitle


\IEEEpubid{\begin{minipage}{\textwidth}\ \\[12pt] \centering
    \IEEEpubidadjcol
  \copyright 2022 IEEE. Personal use of this material is permitted. Permission from IEEE must be obtained for all other uses, in any current or future media, including reprinting/republishing this material for advertising or promotional purposes, creating new collective works, for resale or redistribution to servers or lists, or reuse of any copyrighted component of this work in other works.
\end{minipage}} 


\begin{abstract}
The recent pandemic emergency raised many challenges regarding the countermeasures aimed at containing the virus spread, and constraining the minimum distance between people resulted in one of the most effective strategies. Thus, the implementation of autonomous systems capable of monitoring the so-called \textit{social distance} gained much interest. In this paper, we aim to address this task leveraging a single RGB frame without additional depth sensors. In contrast to existing single-image alternatives failing when ground localization is not available, we rely on single image depth estimation to perceive the 3D structure of the observed scene and estimate the distance between people. During the setup phase, a straightforward calibration procedure, leveraging a scale-aware SLAM algorithm available even on consumer smartphones, allows us to address the scale ambiguity affecting single image depth estimation. We validate our approach through indoor and outdoor images employing a calibrated LiDAR + RGB camera asset. Experimental results highlight that our proposal enables sufficiently reliable estimation of the inter-personal distance to monitor social distancing effectively. This fact confirms that despite its intrinsic ambiguity, if appropriately driven single image depth estimation can be a viable alternative to other depth perception techniques, more expensive and not always feasible in practical applications. 
Our evaluation also highlights that our framework can run reasonably fast and comparably to competitors, even on pure CPU systems. Moreover, its practical deployment on low-power systems is around the corner. 

\end{abstract}
\IEEEpubidadjcol


\IEEEpeerreviewmaketitle

\begin{figure}[t]
    \centering
    \renewcommand{\tabcolsep}{1pt}
    \begin{tabular}{cc}
         \includegraphics[width=0.18\textwidth]{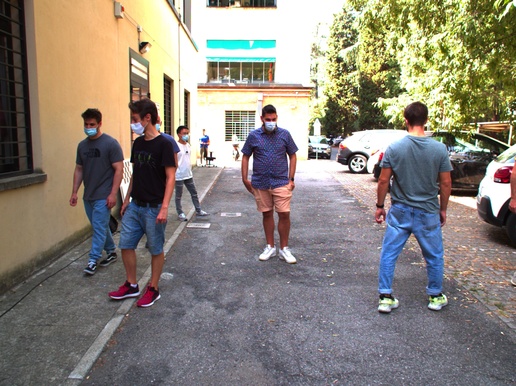} &
         \quad \includegraphics[width=0.18\textwidth]{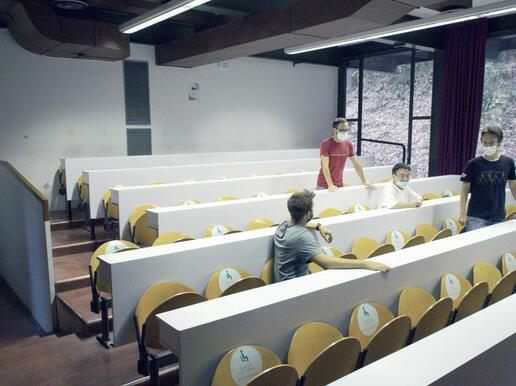} \\
         \includegraphics[width=0.18\textwidth]{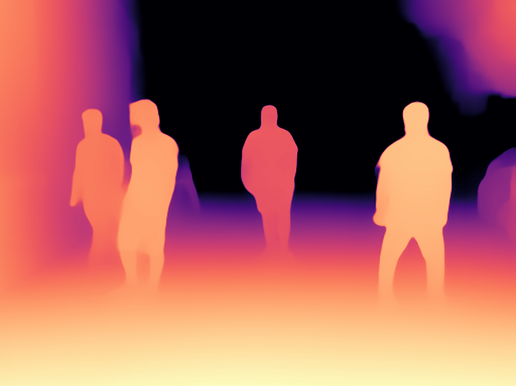} &
         \quad \includegraphics[width=0.18\textwidth]{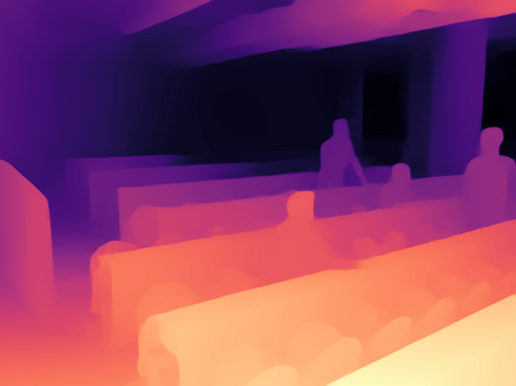} \\
         \includegraphics[width=0.18\textwidth]{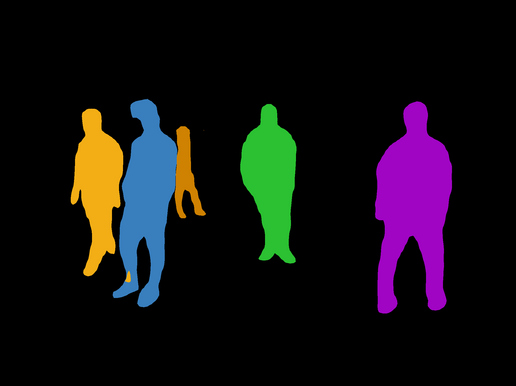} &
         \quad \includegraphics[width=0.18\textwidth]{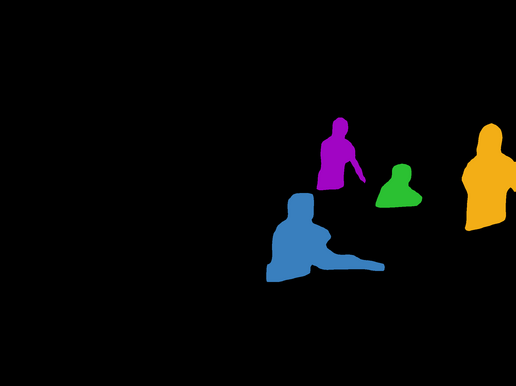}\\
         \includegraphics[width=0.18\textwidth]{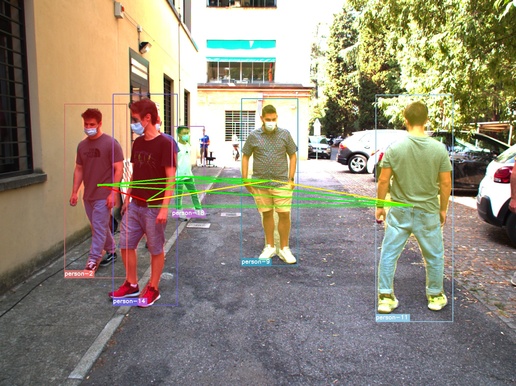} &
         \quad \includegraphics[width=0.18\textwidth]{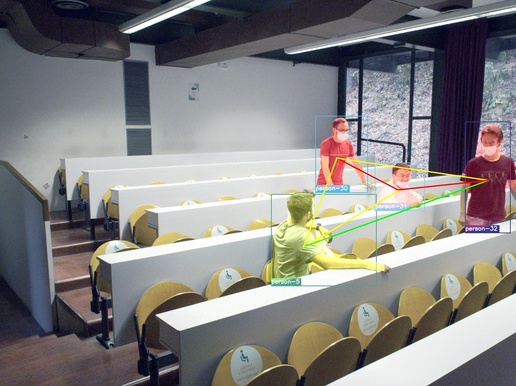} \\
        (a) & (b) \\
    \end{tabular}
    \caption{\textbf{Monitoring social distancing from images.} Images concerning outdoor (a) and indoor (b) environments. From top to bottom: reference RGB image, estimated depth, segmented people, and estimated inter-personal distances, with the color (red, orange, green) of the segment denoting the \textit{class of risk} (dangerous, risky, safe), as explained in Section \ref{detecting_violations}. In contrast to most other methods, our framework reliably estimates the inter-personal distance without constraints about the sensed environment even when dealing with multiple and not visible ground planes (b).}
    \label{fig:teaser}
\end{figure}

\section{Introduction}

\IEEEPARstart{T}{he COVID-19} outbreak started in late 2019 has dramatically impacted humans lives worldwide. In parallel to the development of clinical solutions aimed at both preventing or curing the disease, many good practices and rules have been enforced as countermeasures to avoid infection. Among them, efforts aimed at constraining a higher inter-personal distance between people in shared spaces, known as \textit{social distancing}, proved to be particularly effective at fighting the spread of the virus \cite{khataee2021effects}.
Consequently, the strict observance of such a constraint results critical and thus subject of different studies aimed at developing solutions capable of monitoring the social distancing. 
The underlining problem, consisting of monitoring people's behavior, is not new in the literature \cite{HBA_review} and is at the core of many business intelligence tasks for data analytics  \cite{HBA_people_counting,HBA_video_analytics,HBA_retail,HBA_surveillance,HBA_deep,HBA_edge}. However, it got an urgent and spread interest due to the global emergency. Different technologies \cite{nguyen2020comprehensive1,nguyen2020comprehensive2} allows to implement mechanisms to this end, often leveraging mobile or wearable devices \cite{kumar2021bluetooth,kobayashi2020distancing,rusli2020mysd,johnson2020social,bian2020wearable} to track users' behavior.
\IEEEpubidadjcol

As alternative, solutions based on computer vision \cite{fabbri2020inter,ahamad2020person,shao2021real,Aghaei_2021_WACV,gad2020vision_social_distance}
to determine the \textit{visual} social distance \cite{cristani2020visual} are potentially less intrusive. They do not require explicit commitment by the people to carry mobile/wearable devices and can be adopted using standard cameras deployed for video surveillance purposes. Since extracting this kind of metric information from raw RGB images in the absence of other devices is challenging, most approaches exploit known properties of the scene and context under examination. For instance, it is often common to leverage homography-based localization \cite{fabbri2020inter,Aghaei_2021_WACV,shao2021real,gad2020vision_social_distance} by estimating the ground plane over which people walk or by considering the overlap between detected people bounding boxes \cite{ahamad2020person}.
Although this allows to potentially monitor social distancing from single, uncalibrated images \cite{Aghaei_2021_WACV}, it makes some strong assumptions on the framed scenes. For instance, constraining the task to the necessity of estimating the homography concerning the ground plane makes it impossible to monitor environments where such a plane is not visible. Such a situation frequently occurs, for instance, during many public events such as football games, theater shows or academic lectures, as shown in Fig. \ref{fig:teaser} (b).
Moreover, in order to compensate for the absence of metric knowledge, existing approaches leverage heuristics based on statistics such as, for instance, average people height \cite{Aghaei_2021_WACV}. However, these strategies are not robust in the presence of people whose height deviates significantly from the average or when they are not orthogonally standing on the ground, \eg{} a person sitting. Similarly, using the overlap between bounding boxes \cite{ahamad2020person} is incline to failures due to the perspective.

On the contrary, knowing metric information of the observed scene would allow to directly compute the inter-personal distance between detected actors, ridding of the need for heuristics or statistics -- that could not represent the specific image -- while adding one layer of interpretability to the overall system.
To this aim, \textit{depth perception} can play a crucial role in tackling this problem, as it does for several autonomous systems dealing with navigation and interaction with the surrounding environment. A massive effort in the research community and industry led to accurate depth perception capabilities through different technologies. A rough classification consists of two categories: active and passive strategies.
The former perceives depth by flooding the sensed area with a signal and measuring its behavior. Notable examples are Light Detection And Ranging (LiDAR) \cite{royo2019overview} and Time-of-Flight (ToF) \cite{li2014time} sensors. Despite being accurate, they suffer severe limitations on their working range and are more expensive than standard RGB cameras.
On the other hand, methods belonging to the latter class perceive depth simply from standard images. A notable example is stereo vision \cite{poggi2021synergies}, enabling depth estimation by processing two images sensing the same area from different viewpoints and triangulating correspondences. 

In particular, the passive category has witnessed a paradigm shift in the past few years, originated by introducing learning-based techniques in this field. It enabled to raise the bar dramatically concerning depth estimation accuracy and also to tackle problems inherently ill-posed, such as inferring depth from a single image \cite{Ranftl2020Midas,Ranftl2021MidasTransformer}. 
This approach is highly advantageous since it seamlessly would enable depth perception from the most straightforward and ubiquitous setup made of a single camera. However, on the other hand, this strategy remains intrinsically \textit{scale-ambiguous}, given the ill-posed nature of retrieving depth from a single image, and results in estimating an accurate relative depth, yet not meaningful of metric distances when deployed out of the training domain. Therefore, scale ambiguity is the main limitation when an absolute distance is needed. For this reason, monocular depth perception is seldom deployed in practical applications with such a constraint despite the evident advantages previously mentioned. 

This paper aims to monitor social distancing through a novel framework based on single image depth estimation and a peculiar initialization strategy enabling us to design a robust system for a broad range of situations. As most other existing approaches \cite{fabbri2020inter,gad2020vision_social_distance}, we assume a setup made of a single, static RGB camera coupled with a processing device without any constraint to the sensed environment. However, in contrast to them, we adopt a different approach to tackle social distancing exploiting monocular depth estimation enabling us to avoid any constraint to the sensed environment. We achieve this without depth sensors, relying only on a conventional RGB camera. 
On the one hand, this choice makes our proposal seamlessly adaptable to existing setups made of a single static camera, such as the countless surveillance cameras spread worldwide, without any modification. Moreover, it would also allow low budget installation costs when facing new setups, and at the same time, overcoming the limitations of existing single-camera solutions \cite{fabbri2020inter,Aghaei_2021_WACV}.
On the other hand, for the same reasons outlined before, neglecting the deployment of alternative depth sensing techniques such as active sensing technologies like LiDAR or stereo vision makes social distancing estimation much more challenging and only partially explored in the literature. Furthermore, like in most existing setups, we assume only the availability of a single static RGB camera during the deployment. Thus, we cannot rely on passive techniques requiring a moving camera such as structure-from-motion or Simultaneous Localization And Mapping (SLAM) to infer depth at runtime. Indeed, as outlined so far, it is highly relevant these days to enable reliable social distancing from a single static camera seamlessly, neither with expensive/cumbersome depth sensors nor the limitations of existing monocular approaches. Purposely, we only assume that, during system installation, an operator can use for a few seconds a standard smartphone to infer a sparse set of scaled 3D points -- leveraging a scale-aware SLAM approach \cite{arcore,arkit} -- overlapping with the area sensed by the fixed camera. Once remapped to the reference system of the fixed camera, such sparse points will be used to scale the relative monocular depth to absolute measures facing the previously mentioned issues. It is worth stressing that the procedure outlined is fast and is required only once upon system initialization. Indeed, such a procedure shall be repeated only whether a large portion of the 3D structure of the background changes.
At execution time, our system leverages an off-the-shelf and pre-trained single image depth estimator \cite{Ranftl2020Midas,Ranftl2021MidasTransformer,poggi2018pydnet} to estimate relative depth maps that we rescale according to the few known 3D points obtained during the setup stage. This strategy allows for obtaining dense, metric depth measurements for the entire scene. Finally, people are segmented through an instance segmentation network \cite{he2017mask} and their inter-personal distance is estimated according to their estimated 3D position in space. 
To assess the effectiveness of our framework, we collect a set of indoor and outdoor RGBD examples using a previously calibrated LiDAR + RGB camera asset used as a reference to measure inter-personal distances. Experimental results confirm that our framework effectively delivers sufficiently accurate metric measurements and thus can be deployed to monitor social distancing without constraints to the sensed scene, such as a visible ground plane, needed by other monocular approaches. 

Experiments on real datasets support the following claims:
\begin{itemize}
    \item We prove that monocular depth estimation, an emerging research topic in computer vision, can effectively tackle non-critical applications requiring metric measurements such as social-distancing monitoring.
    \item Information about the unknown scale of the scene, required to retrieve metric predictions from the output of a monocular depth estimation network, can be sourced through a quick and straightforward offline initialization, requiring for a few seconds only a standard consumer device like a smartphone. 
    \item Unlike other single-camera approaches not exploiting monocular depth estimation, our method is robust against well-known issues such as the need for a visible ground plane. Moreover, when the ground plane is visible, it performs comparably or better \cite{Aghaei_2021_WACV} than methods strictly requiring it. 
\end{itemize}

\section{Related work}

We briefly review the literature relevant to our approach, including social distancing monitoring, depth estimation and single image depth estimation.

\subsection{Social distance monitoring} Measuring the inter-personal distance is a crucial task to monitor social distancing and thus prevent the spread of the COVID-19 pandemic. Several approaches arose in the last years \cite{nguyen2020comprehensive1,nguyen2020comprehensive2}, exploiting different technologies.
A popular choice consists of using dedicated devices sharing information utilizing a network \cite{kumar2021bluetooth,kobayashi2020distancing,rusli2020mysd,johnson2020social,bian2020wearable}. However, these solutions require an ad-hoc infrastructure and a known communication protocol between the users and the backend. Unfortunately, this constraint cannot be enforced in many circumstances, such as in crowded places (\eg{} stations), and require a direct collaboration of the monitored users (which have, for instance, to install an app on their mobile phone or to wear a custom Bluetooth device). Conversely, passive technologies -- and specifically vision-based solutions -- enable to monitor the distance among sensed people without a direct and voluntary collaboration from the users, potentially exploiting already available infrastructures such as surveillance cameras.
Since the system has to work in new environments, existing strategies \cite{gad2020vision_social_distance,fabbri2020inter,Aghaei_2021_WACV} rely on a preliminary calibration phase. Specifically, \cite{gad2020vision_social_distance,fabbri2020inter} explicitly leverage the pre-determined homography between camera and ground plane to determine the 3D position on the floor of each person locating his/her feet. Additionally, \cite{fabbri2020inter} tries to infer the feet position even when they are occluded through a deep neural network. Differently, \cite{Aghaei_2021_WACV} given an approximated knowledge of the orientation of the camera with respect to the ground plane, by assuming a fixed parameter of the human body (specifically, the constant height of 1.70 m) inferred by a deep network, detects social distancing violations by leveraging the heuristically determined torso length of each person.
On the contrary, with the same setup, the proposed method aims to solve the task through a completely different strategy based on direct dense depth estimation. It requires an initial calibration phase only to obtain control points in the static background scene -- not constrained to the ground plane -- necessary to retrieve the metric scale of the scene. Moreover, our method has no additional constraints, such as, for instance, the need for a non-occluded contact point between each person to track and the plane at test time. 

\subsection{Depth Estimation} Active sensors, such as LiDAR \cite{royo2019overview}, structured lights or ToF \cite{li2014time}, can source accurate depth measurements. However, they suffer the problems mentioned above and are not widespread on every device due to cost, energy consumption and space limitations. Thus, obtaining depth from cameras represents a popular choice, especially in mobile devices \cite{valentin2018mobile}, enabling AR applications on-board \cite{du2020depthlab}. Structure from Motion (SfM) \cite{schoenberger2016sfm,engel2017direct} and SLAM \cite{schops2017large,campos2020orbslam3} provide depth measurements as well as the poses of the camera, but suffer in dynamic scenes.
On the contrary, in this paper, we rely on a single image depth estimation network, making our system appropriate for a large set of devices (potentially, any standard RGB camera). Moreover, it can work properly even in the case of a static camera at test time, which is a standard setup in surveillance applications.  

\subsection{Single Image Depth Estimation} Obtaining the scene's depth given a single view is an extremely challenging problem, and deep learning seems to be a promising strategy to tackle it. Supervised \cite{eigen2014depth, fu2018deep,bhat2021adabins} approaches require depth information usually sourced by means of active sensors such as LiDAR or Kinect devices, self-supervised strategies require stereo pairs or monocular video sequences at training time \cite{godard2017monodepth,tosi2019learning},  \cite{zhou2017unsupervised,godard2019monodepth2,guizilini20203d}. These training strategies allow obtaining competitive results with images similar to the training domain. However, they suffer when the testing environment is different from the training one (\eg{}, when models are trained outdoor but tested indoor). To overcome this issue, some solutions towards the direction of strong generalization capabilities have been proposed \cite{Ranftl2020Midas,Ranftl2021MidasTransformer,aleotti2020real,xian2020structure,yin2021learning,MegaDepthLi18}, alleviating the need for retraining the model when the environment changes. Finally, compact networks have been designed as well \cite{poggi2018pydnet,wofk2019fastdepth,Ranftl2020Midas} to obtain depth maps out of single images with high fps even on consumer devices, as mobile phones. Finally, it is worth noticing that many monocular networks infer \textit{inverse depths} \cite{godard2019monodepth2, Ranftl2020Midas, Ranftl2021MidasTransformer} or \textit{disparities} \cite{godard2017monodepth,tosi2019learning}, and not metric depth values directly.

\subsection{Instance segmentation}
Instance segmentation \cite{he2017mask}, \ie{} assigning unique labels to pixels belonging to any single object in the scene, is a popular problem in scene understanding at the intersection between semantic segmentation \cite{chen2017rethinking,takikawa2019gated} and object detection \cite{redmon2018yolov3,lin2017focal}.
State-of-the-art solutions \cite{he2017mask,Chen2018_deeplab,wang2020solo,liu2021Swin,wu2019detectron2} employ complex architectures, running at few FPS on modern GPUs. Nonetheless, compact and fast models have been recently proposed \cite{bolya2019yolact,cao2020sipmask,chen2020blendmask,liu2021yolactedge,du2021real}. In this work, we leverage the instance segmentation task using \cite{liu2021yolactedge} to identify people, assign them a unique label and filter control points occluded by them.

\begin{figure}
  \centering
  \includegraphics[width=0.8\linewidth]{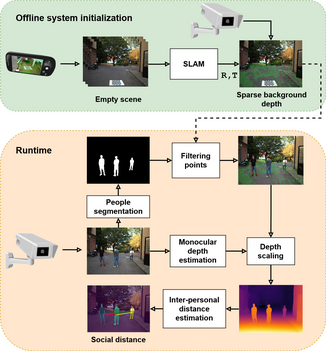}
  \caption{\textbf{Illustration of the proposed pipeline.} In an offline calibration phase (green box), we source control points in the environment employing off-the-shelf strategies. Then, when a new frame is available at runtime (orange box), we rescale the depth provided by a monocular network thanks to the available control points in the background and segment people using an instance segmentation network. Once each person's depth and mask in the scene are predicted, we can estimate the inter-personal distance.}
  \label{fig:pipeline}
\end{figure}

\section{Proposed method}

This section describes our framework, providing a detailed description of each of its components for social distance monitoring with a single static camera. Before its deployment, it requires a quick offline calibration carried out with a standard smartphone, equipped with a single camera in our experiments. Once completed, the framework provides a metric estimation of the distance between people using only the fixed camera. The complete pipeline is described in Fig. \ref{fig:pipeline}.

\subsection{Offline system initialization}\label{offline_calib}

In the initialization phase, executed only once during the first installation, a straightforward calibration procedure is needed to obtain a sparse 3D structure of the scene. This task can be accomplished in a few seconds using any device capable of inferring depth at a reasonably known scale, such as an active sensor, a stereo vision system or other techniques like SLAM. To minimize as much as possible installation requirements, in our experiments, we accomplish this task relying on the SLAM capability provided by the ARCore framework for augmented reality available for Android devices. Nonetheless, using the ARKit framework for iOS would be equivalent. These frameworks rely on an Inertial Measurement Unit (IMU) measurements and at least a single image stream to infer a sparse map of the sensed environment at a known scale. 
Moreover, it is worth noticing that the control points could be sourced even with more accurate and expensive devices -- such as LiDARs -- or with different strategies such as a full-featured SLAM system like ORB-SLAM3 \cite{campos2020orbslam3} coupled with IMU measurements. Finally, we point out that augmented reality frameworks can seamlessly take advantage of additional setups like stereo or active sensors often available in smartphones or tablets. Nonetheless, we stick to the most constrained setup using a single camera for our experiments to reduce as much as possible the installation requirements. Specifically, the \textit{Raw Depth API} \cite{ARCore_Raw_Depth_API} provided by ARCore enables us to obtain the sparse 3D structure of the target area within a range of [0-8] meters, providing also a confidence estimation for such data.

\begin{figure}[t]
  \centering
  \includegraphics[width=0.8\linewidth]{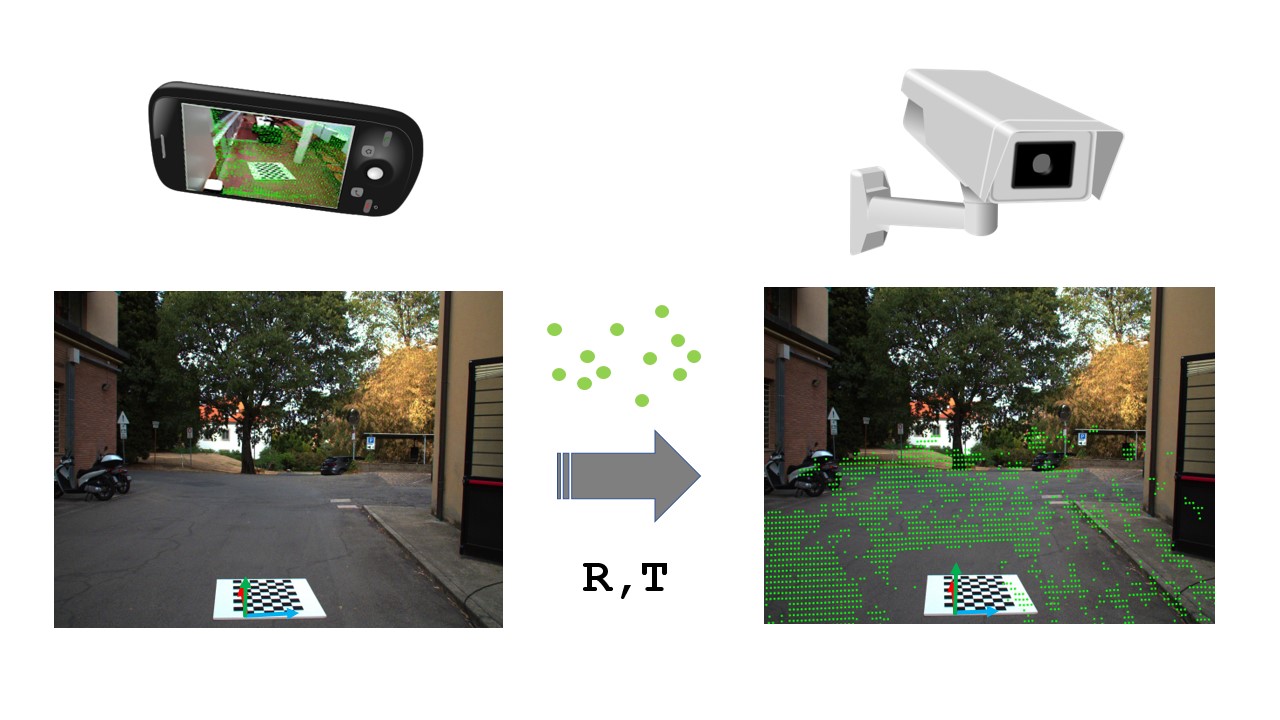}
  \caption{\textbf{System initialization.} The pointcloud inferred by the smartphone on the left is remapped into the fixed camera reference system on the right by knowing the relative camera pose R,T between the two devices. Although four coplanar yet not collinear points at a known relative position would suffice to accomplish this task, for better clarity, we use in this figure a chessboard.}
  \label{fig:initialization}
\end{figure}

\begin{figure*}[ht]
    \centering
    \scalebox{0.9}{
    \begin{tabular}{cccc}
        \includegraphics[height=0.14\textwidth]{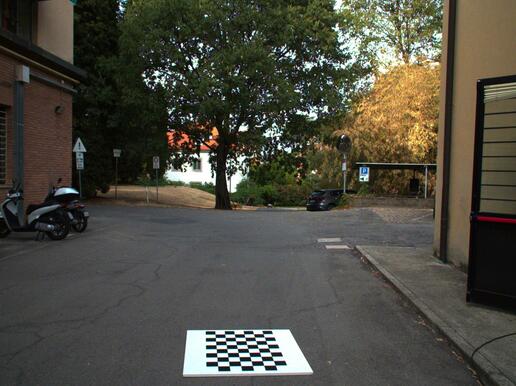} &
        \includegraphics[height=0.14\textwidth]{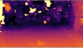} &
        \includegraphics[height=0.14\textwidth]{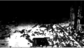} & 
        \includegraphics[height=0.14\textwidth]{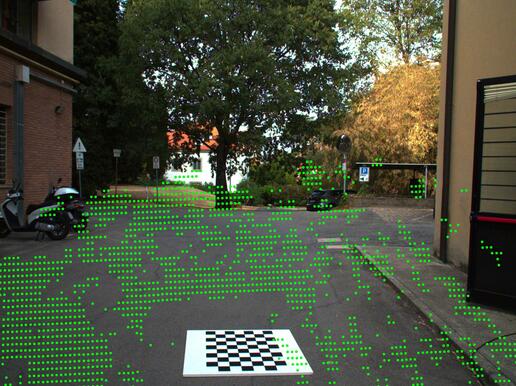} \\
        RGB ($2064\times1544$) & Depth ($160\times90$) & Confidence ($160\times90$) & Control Points ($2064\times1544$)
    \end{tabular}
    }
    \caption{\textbf{Example of control points sourcing using ARCore.} We adopt the off-the-shelf SLAM framework provided by ARCore on Android devices to source control points. The estimated depth is filtered by exploiting the native pixel-wise confidence score (higher the confidence, lighter the color) provided by the same framework. Notice that the depth map generated by ARCore is way smaller than the fixed camera image, as well as of the image acquired by the smartphone. Thus, at the end of the projection onto the image of the fixed camera, only just a few sparse measures are available as control points.}
    \label{fig:arcore_qualitative}
\end{figure*}

As depicted in Fig. \ref{fig:initialization}, at system installation, the operator selects or places at least four coplanar yet not collinear points in the sensed scene at a known relative position and records the scene moving a smartphone or a tablet in the target (although not strictly required) empty area. Then, leveraging the position of such points in both the mobile camera image and in the fixed camera, it is possible to infer the relative position between them by solving the \textit{Perspective-n-Point} problem \cite{Lu_2018}. This procedure can be either automated using a well-known pattern such as the chessboard in Fig. \ref{fig:arcore_qualitative} or manual selecting at least four known points in the scene. It is worth observing that this procedure can be carried out even by a not-specialized operator with a consumer smartphone through a simple guided step-by-step wizard.
Once the camera pose R, T between the two cameras is known, we can move depth measurements $D_\text{SLAM}$ into the fixed camera reference system obtaining $D_\text{cam}$ as
\begin{equation}
    D_\text{cam} \sim (R|T)D_\text{SLAM}
\end{equation}
The newly obtained depth values $D_\text{cam}$ are projected from $p_\text{SLAM}$ coordinates in the mobile device camera to $p_\text{cam}$ pixel coordinates in the fixed camera view
\begin{equation}
    p_\text{cam} \sim K (R|T)D_\text{SLAM}K^{-1}p_\text{SLAM}
\end{equation}
and used as control points to perform the depth rescaling at runtime. In any case, we exploit the pixel-wise native confidence estimation of ARCore to filter out the unreliable depth values, selecting only the control points with the maximum confidence score. Fig. \ref{fig:arcore_qualitative} shows an example of control point projection over the fixed camera. Finally, it is worth noticing that control points must be sourced on fixed objects that would be available (if not occluded) even at test time, such as points on walls, furniture or trees. So, the sensed area should not contain people in the initialization phase and keep the same 3D background structure at execution time.

\subsection{People segmentation}\label{people_segmentation}

For each new image acquired by the fixed camera, we identify at first people silhouettes through an istance segmentation network. For this purpose, we rely on YolactEdge \cite{liu2021yolactedge}, a fast and lightweight instance segmentation network. Specifically, starting from the pre-trained model made available by the authors, we specialize it for people through a fine-tuning for 24 epochs on a subset of the COCO 2017 dataset \cite{lin2014microsoft} with images mainly containing people. The silhouettes obtained by the segmentation network are used for two purposes: i) detect people to compute their inter-personal distances and ii) remove from the set of depth points inferred in the initialization stage those occluded by people, in order to scale the output of the monocular depth estimation network appropriately. Concerning the latter point, the underlying assumption is that the 3D structure of the background, where the 3D ground control points lay, does not change at runtime. To this aim, we remove people after the segmentation step. It is worth observing that, in case of considerable modifications to the underlying 3D background structure of the scene with respect to the one acquired during the system setup, a new initialization would be necessary. This requirement might occur in two extreme cases: either when most of the 3D structure of the background or the camera position has changed. In these cases, a whole initialization phase requires just a few minutes.

\subsection{Monocular depth perception and scaling}

For the reasons previously stated, we decided to rely on pre-trained networks for depth estimation without performing any fine-tuning in the target domain. Consequently, we focus our attention on networks capable of generalizing well to heterogeneous environments thanks to supervised training on large collections of data. Although it does not allow us to retrieve scaled maps from the raw output of the monocular depth network, this strategy avoids cumbersome and time-consuming data collection required to fine-tune the network in each target domain. Moreover, it seamlessly allows replacing the depth estimation backbone with improved/newly released ones. Specifically, in our experiments, we rely on the following pre-trained monocular depth estimation networks. 

\textbf{MIDAS.} Proposed by Ranftl \textit{et al} \cite{Ranftl2020Midas}, MIDAS is a convolutional neural network model trained to predict inverse depth maps for a large set of different data, such as 3D movies or stereo videos. The authors provide two models, characterized by different encoders and thus by different complexity: the large model, counting about 105 M parameters, and the small one, with 21 M parameters, referred to as MIDAS and MIDAS small. While the latter runs at a high framerate even on mobile devices (MIDAS small v2.1 runs at 30 FPS on an Apple iPhone 11, as reported in the official GitHub repository), the former is more accurate. In our experiments, we use both models with version ID v2.1.  

\textbf{DPT.} The Dense Prediction Transformer (DPT) \cite{Ranftl2021MidasTransformer} represents the state-of-the-art solution for monocular depth estimation in the wild. It is based on a Visual Transformer (ViT \cite{dosovitskiy2020image}) backbone that takes a set of flattened embeddings (tokens) from non-overlapping patches of the image and process them through a set of transformer layers. Then, a convolutional decoder reassembles the tokens into an image-like feature representation at different resolutions that are progressively fused. Finally, a dedicated head predicts the task-specific output at the original input resolution. Eventually, the tokens can be source using a CNN backbone: in this case, the model is called ViT-Hybrid, and exploits a ResNet50 \cite{he2016deep}. We rely on the official DPT and DPT-Hybrid models released by the authors trained for monocular depth estimation task, \ie{} according to the procedure outlined in \cite{Ranftl2020Midas}. DPT and DPT-Hybrid count nearly 344 and 123 M parameters, respectively.

\textbf{PyD-Net.} Proposed in \cite{poggi2018pydnet}, PyD-Net is a lightweight CNN network made of an encoder and multiple decoders, each of them in charge of predicting an inverse depth at a specific resolution. This strategy allows stopping the execution at a lower resolution when the model runs on low-power devices, preserving computational resources. The network counts just 1.9 M parameters. In this work, we adopt the variant trained to be robust in the wild proposed in \cite{aleotti2020real}. 

We evaluate the performance of the proposed system leveraging these pre-trained monocular networks. However, despite their excellent generalization capability to different scene contents, they provide only relative depth measurements and a scaling process is mandatory to obtain metric distances. Purposely, we rely on the sparse background depth points acquired in the initialization stage and filtered as previously outlined. We exploit such points to scale the output of the network at each frame as follows. Usually, monocular depth networks output the relative inverse depth for each pixel of the image; thus is mandatory to bring the sparse background control points into such domain. Given such background depth points, obtained as detailed in section \ref{offline_calib}, and the corresponding relative inverse depth points inferred by the monocular depth network, we pre-compute the inverse depth of the former and then, assuming a linear data distribution \cite{Ranftl2020Midas}, we find the best fit between such correspondences through a linear regression to obtain the best offset $\alpha$ and slope $\beta$. To perform the linear regression task, we use the least-squares method as in \cite{Ranftl2020Midas}. Specifically, considering each relative inverse depth point $x_i$ with a valid inverse metric correspondence $y_i$ available, we obtain two scalars $\alpha$ and $\beta$ as follows, where $n$ is the total number of available depth points with a valid correspondence:

\begin{align}
    \label{eq:linear_regression}
    \begin{bmatrix}\alpha \\ \beta\end{bmatrix}
    =
    \begin{bmatrix}
    n & \sum_{i=1}^n{x_i} \\
    \sum_{i=1}^n{x_i} & \sum_{i=1}^n{x_i^2}
    \end{bmatrix}^{-1}
    \begin{bmatrix}
    \sum_{i=1}^n{y_i} \\ \sum_{i=1}^n{x_i y_i}
    \end{bmatrix}
\end{align}

Then, we scale according to $y = \alpha + \beta x$ the whole relative inverse depth map inferred initially by the network. Finally, we move back to depth domain and obtain metric distances. 

It is worth noticing that theoretically, only a set of two correspondences would suffice to obtain the $\alpha$ and $\beta$ to scale the relative depth map. Nonetheless, using more points improves robustness to outliers in at least one of the two sets. Since the output of the monocular depth network is entirely dense, the number of points for the regression task coincides with those acquired in the initialization stage, typically a few thousand, surviving the previous phase.

\subsection{Computing inter-personal distance}\label{inter-personal-distance}

Given the dense scaled depth map and the silhouette of each person obtained through an instance segmentation network such as \cite{liu2021yolactedge} or \cite{wu2019detectron2}, the final step aims at computing people's distance. Although, in theory, the minimum distance between two people is the closest distance between two points belonging to each one, deploying this strategy for social distancing would be computationally expensive and prone to errors occurring primarily due to depth estimation inaccuracy. Consequently, we rely on a more robust yet approximated strategy inferring the distance between the \textit{centroid} representing a point in space for each person in the sensed scene.
Specifically, given the set of pixel coordinates assigned to a specific person $\Omega$ (defined by the people segmentation mask obtained as described in Section \ref{people_segmentation}), we compute the $(u,v)$ coordinates of the centroid $\mathcal{C}$ in the image as follows:
\begin{equation}
    \begin{array}{c}
    \mathcal{C}_u =\frac{1}{|\Omega|} \sum_{p\in\Omega} u_p  \\
    \mathcal{C}_v =\frac{1}{|\Omega|} \sum_{p\in\Omega} v_p  \\
    \end{array}
\end{equation}

Unfortunately, the computed $\mathcal{C}$ may not match with any of the points in $\Omega$ (\eg{} due to occlusions, like a person sitting behind a bench). In such cases, we use the closest point contained in $\Omega$ to $\mathcal{C}$ as the centroid. The depth of $\mathcal{C}$ is sourced from the scaled depth map, allowing us to back-project $\mathcal{C}$ in the 3D space. Finally, we obtain the inter-personal distance as the Euclidean distance between each pair of centroids.

\begin{figure}[t]
    \centering
    \scalebox{0.5}
    {
    \begin{tabular}{ccc}
    
         \includegraphics[width=0.3\textwidth]{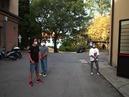} &
         \includegraphics[width=0.3\textwidth]{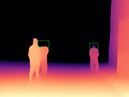}&
         \includegraphics[width=0.3\textwidth]{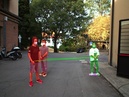}\\
         RGB & Monocular Depth & Final Result
    \end{tabular}
    }
    \caption{\textbf{Example of monocular depth failure.} The head of the person is missed by the network. Nonetheless, the centroid is found on the body.}
    \label{fig:monocular_failure}
\end{figure}

This strategy, even if simple, is reliable against potential errors in monocular depth estimation. Specifically, as depicted in Fig. \ref{fig:monocular_failure}, some portions of a person, such as the head, are critical and challenging to predict correctly for many current monocular networks. In contrast, the segmentation network identifies them reliably. Thus, to increase robustness, our strategy segments the person in the 2D space and determines the 3D position of its centroid.

\section{Experimental results}

In the following section, we aim at evaluating the proposed strategy in real environments. Since open-source data sets with accurate depth values for testing purposes are not available, we collect our own set of images coupled with depth information to assess the performance of the proposed approach in indoor and outdoor settings. Specifically, we rely on an asset composed by a Livox Mid-70 LiDAR sensor and an RGB camera (FLIR BFS-U3-32S4C-C) previously calibrated and registered using a proper pattern to obtain their relative pose. It is worth noticing that this further calibration is needed only to prepare our evaluation benchmark and not for system deployment -- which requires only the offline calibration presented in Section \ref{offline_calib}. Moreover, the control points are acquired using a Google Pixel2 XL and a OnePlus 6 smartphone to stress that a standard commercial device suffices in setting up the system. The point cloud obtained from the LiDAR is projected over the RGB frame leading to about 6000 depth points. However, leveraging the non-repetitive scanning technology of this specific LiDAR sensor, we can accumulate multiple point clouds of the same static scene to obtain much denser depth maps. Assuming a static scene for N consecutive LiDAR acquisitions, we can collect about 120000 points for each RGB-D example. The number of frames N is fixed to 20, and the RGB image linked to the depth data is the last one acquired by the camera. All the devices involved are not synchronized, but this is not an issue due to the constraints imposed by the previously described accumulation process, which can deliver high-quality depth maps used as a reference.

In the reminder, we first evaluate the accuracy of the depth points acquired through the SLAM module available in ARCore. Then, we describe the sequences acquired to assess the effectiveness of our strategy. Lastly, according to different metrics, we evaluate the effectiveness of monitoring social distancing employing the output of monocular depth networks scaled using background control points as outlined in our proposal. Purposely, we consider in our evaluation state-of-the-art monocular depth estimation networks capable of generalizing to different and unpredictable environments without requiring additional fine-tuning in the target scene. Consequently, we include the following networks using the authors' trained models: MIDAS and MIDAS Small \cite{Ranftl2020Midas}, DPT and DPT-Hybrid \cite{Ranftl2021MidasTransformer}, and PyD-Net \cite{poggi2018pydnet} trained as in \cite{aleotti2020real}.

\begin{figure}[t]
    \centering
    \renewcommand{\tabcolsep}{1pt}
    \scalebox{0.75}{
    \begin{tabular}{ccccc}

        ARCore & ARCore & & &  ARCore \\
        Cam & depth & Cam & LiDAR &  depth $\rightarrow$ Cam \\
        \includegraphics[align=c,width=0.12\textwidth]{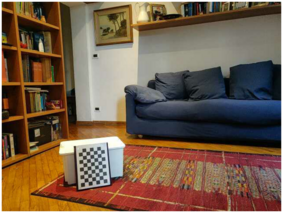}                & 
         \includegraphics[align=c,width=0.12\textwidth]{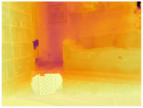}    &
        \includegraphics[align=c,width=0.12\textwidth]{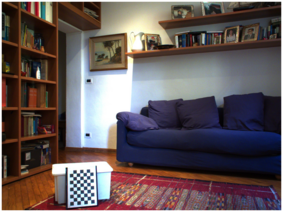}                &
        \includegraphics[align=c,width=0.12\textwidth]{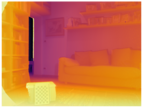} &
        \includegraphics[align=c,width=0.12\textwidth]{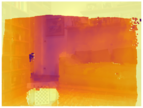}  \\

        \includegraphics[align=c,width=0.12\textwidth]{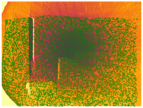}   &
        \includegraphics[align=c,width=0.12\textwidth]{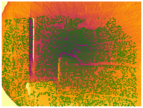}  & 
        \includegraphics[align=c,width=0.12\textwidth]{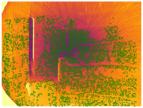}  & 
        \includegraphics[align=c,width=0.12\textwidth]{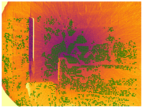}  & 
        \includegraphics[align=c,width=0.12\textwidth]{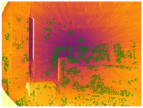}  \\
        Th = 0 & Th = 0.2 & Th = 0.4 & Th = 0.6 & Th = 0.8                \\
        \multicolumn{5}{c}{
            \includegraphics[align=c,width=0.60\textwidth]{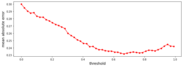}
        }
    \end{tabular}
    }
    \caption{\textbf{Evaluation of depth data inferred with the ARCore raw depth APIs.} The first row depicts, from left to right: the image acquired with the smartphone, the depth map inferred with ARCore, the image acquired by the static camera, the depth map acquired by the LiDAR registered with the static camera, the ARCore points projected onto the fixed camera point of view. In the first row, for visualization purposes, we dilate the sparse ARCore and LiDAR depth maps to densify them and match the highest cameras resolution. From the point of view of the static camera, the second row shows the LiDAR points and the sparse points (green) obtained through ARCore for different confidence thresholds, from 0.0 to 0.8. The graph at the bottom reports the MAE of ARCore depth point wrt the LiDAR points for confidence threshold ranging from 0.0 to 1.0.}
    \label{fig:ARCore_evaluation}
\end{figure}

\subsection{Evaluation of control points accuracy}\label{evaluation_sfm}

Since our proposal relies on the background control points to scale the depth maps inferred by monocular networks, we start by evaluating the accuracy of the depth obtained by the SLAM module available in ARCore. For this purpose, we scanned the indoor environment depicted in Fig. \ref{fig:ARCore_evaluation} with a OnePlus 6 smartphone to compute the ARCore depth map (we report the image acquired and the corresponding depth map in the two leftmost positions at the top of the figure).  Moreover, we also framed the same scene with the previously described RGB-D asset but accumulating about 500 frames from the LiDAR sensor leading to a sparse depth map with a 95\% density to better match the control points. The static camera and the LiDAR dense depth map are shown in the third and fourth positions of the first row of the Fig. \ref{fig:ARCore_evaluation}. Then, we determined the pose of the smartphone wrt the fixed camera frame and projected the ARCore depth points onto the latter. As shown in the rightmost figure at the top, we can notice how the 3D structure of the scene inferred by ARCore is only partially overlapping with the actual image, pointing out not slight inconsistencies in the depth map. Its amount and impact on the scaling process will be analyzed more in detail next. 

In the second row, we show the LiDAR depth map from the fixed camera point of view and, filtering by different ARCore confidence threshold levels, the points inferred by ARCore in green. We can notice that even not applying the confidence filtering (th=0), the number of points inferrable with ARCore (bounded to 160$\times$90) is much lower than those of the LiDAR but accumulating multiple scans of a static scene. However, without such accumulation, the number of points provided by the Livox LiDAR in a single scan would be much lower (about 6000). Not unexpectedly, we can notice that the higher the confidence, the lower the number of ARCore points surviving. Nonetheless, more interestingly, even setting a high confidence threshold (eg, th=0.8), the number of estimated confidence points is not negligible and quite spread across the whole scene in this image.
Finally, by looking at the graph at the bottom of Fig. \ref{fig:ARCore_evaluation} where we plot the MAE between the overlapping ARCore and LiDAR depth points for different threshold levels, we can observe that the error range from 30 to 23 cm in this same scene. Such a magnitude is comparable to other measurements carried out on other scenes. Moreover, we can also notice that the error decreases almost monotonically until the confidence threshold is around 0.6 and then oscillates by about 1 cm. In other evaluations, we noticed a similar behavior but better accuracy by increasing the threshold further. From this analysis and the fact that only a few yet reliable points are needed to perform the scaling process, assuming that we cannot perform this evaluation in a practical application, we conservatively set the highest threshold of 1.0 in the experiments reported in the subsequent experiments.

\begin{figure*}[t]
    \centering
    \renewcommand{\tabcolsep}{1pt}
    \scalebox{0.75}{
    \begin{tabular}{ccccccc}
         &  \textbf{Outdoor} & \textbf{Outdoor} & \textbf{Outdoor} & \textbf{Indoor} & \textbf{Indoor} & \textbf{Indoor}  \\
        \rotatebox{90} {\textbf{RGB}} & \includegraphics[align=c,width=0.18\textwidth]{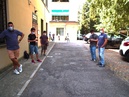} & \includegraphics[align=c,width=0.18\textwidth]{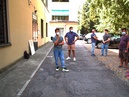} &  \includegraphics[align=c,width=0.18\textwidth]{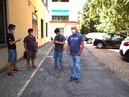} & 
        \includegraphics[align=c,width=0.18\textwidth]{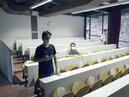} &
        \includegraphics[align=c,width=0.18\textwidth]{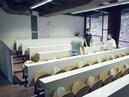} & 
        \includegraphics[align=c,width=0.18\textwidth]{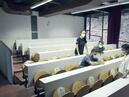} \\
        \rotatebox{90}{\textbf{Depth}} &  \includegraphics[align=c,width=0.18\textwidth]{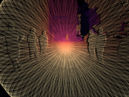} & \includegraphics[align=c,width=0.18\textwidth]{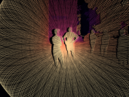}& \includegraphics[align=c,width=0.18\textwidth]{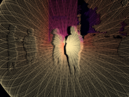} & 
        \includegraphics[align=c,width=0.18\textwidth]{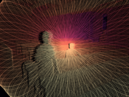} &
        \includegraphics[align=c,width=0.18\textwidth]{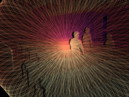} &
        \includegraphics[align=c,width=0.18\textwidth]{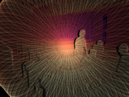} \\

    \end{tabular}
    }
    \caption{\textbf{Examples from the collected dataset.} In the first row, we report three outdoor and three indoor images framing still people. The corresponding LiDAR scans obtained through a Livox Mid-70, accumulating depth points for 20 frames, are reported in the second row.}
    \label{fig:dataset}
\end{figure*}

 \begin{table*}[t]
 \centering
 \scalebox{0.75}{
 \begin{tabular}{ccc}
 \begin{tabular}{l|cc|cc}
    \multicolumn{1}{c}{} & \multicolumn{2}{c}{Any range} & \multicolumn{2}{c}{$<$3 meters} \\
    \hline
    Model & Outdoor & Indoor & Outdoor & Indoor \\ 
    \hline
    PyD-Net \cite{poggi2018pydnet} & 0.44 & 0.76 & 0.33 & 0.82\\
    MIDAS \cite{Ranftl2020Midas} & 0.58 & 0.42 & 0.39 & 0.27 \\
    MIDAS small \cite{Ranftl2020Midas} & 0.69 & 1.09 & 0.52 & 0.97 \\
    DPT-Hybrid \cite{Ranftl2021MidasTransformer} & 0.44 & 0.73 & 0.27 & 0.46\\
    DPT \cite{Ranftl2021MidasTransformer} & 0.48 & 0.65 & 0.31 & 0.39\\
    \hline
 \end{tabular}
&
  \begin{tabular}{l|cc|cc}
    \multicolumn{1}{c}{} & \multicolumn{2}{c}{Any range} & \multicolumn{2}{c}{$<$3 meters} \\
    \hline
    Model & Outdoor & Indoor & Outdoor & Indoor \\ 
    \hline
    PyD-Net \cite{poggi2018pydnet} & 0.58 & 0.51 & 0.47 & 0.34 \\
    MIDAS \cite{Ranftl2020Midas} & 0.54 & 0.43 & 0.32 & 0.32 \\
    MIDAS small \cite{Ranftl2020Midas} & 0.77 & 0.62 & 0.67 & 0.49 \\
    DPT-Hybrid \cite{Ranftl2021MidasTransformer} & 0.29 & 0.39 & 0.24 & 0.23 \\
    DPT \cite{Ranftl2021MidasTransformer} & 0.33 & 0.44 & 0.24 & 0.30 \\
    \hline
 \end{tabular}
&
  \begin{tabular}{l|cc|cc}
    \multicolumn{1}{c}{} & \multicolumn{2}{c}{Any range} & \multicolumn{2}{c}{$<$3 meters} \\
    \hline
    Model & Outdoor & Indoor & Outdoor & Indoor \\ 
    \hline
    PyD-Net \cite{poggi2018pydnet} & 0.42 & 0.53 & 0.33 & 0.34 \\
    MIDAS \cite{Ranftl2020Midas} & 0.44 & 0.43 & 0.35 & 0.31 \\
    MIDAS small \cite{Ranftl2020Midas} & 0.60 & 0.61 & 0.53 & 0.48 \\
    DPT-Hybrid \cite{Ranftl2021MidasTransformer} & 0.34 & 0.40 & 0.25 & 0.23 \\
    DPT \cite{Ranftl2021MidasTransformer} & 0.37 & 0.44 & 0.31 & 0.30 \\
    \hline
 \end{tabular}
 \\
 \bfseries (a) ARCore points & \bfseries (b) Livox points & \bfseries (c) Livox points ($< 10$ meters) \\
 \\
 \end{tabular}
 }
 \caption{\textbf{Evaluation of Inter-Personal distances.} system scaled with (a) ARCore points using confidence threshold 1, (b) unconstrained LiDAR measurements, and (c) LiDAR measurements below 10 meters. The inter-personal distance predictions are compared to the reference depth provided by the LiDAR, averaged over all pairs and images. We report results in terms of MAE for the indoor and outdoor sequences, considering unconstrained inter-personal distances and below 3 meters. }
 \label{tab:distance_evaluation}
 \end{table*}

\subsection{Dataset}\label{dataset}
We collected two sequences, framing indoor and outdoor domains, for a total of 83 frames with accurate depth labels. Fig. \ref{fig:dataset} depicts frames sampled from each sequence.

In the following, we provide additional details for each collected sequence. We refer to control points as those scene points with an associated depth value obtained through the ARCore SLAM framework, which has a confidence score, estimated by ARCore, equal to one. 

\textbf{Outdoor sequence}
The scene, counting 46 frames, depicts six people walking in a driveway with a poorly textured yellow wall at the left and parked cars at the right. The scene contains 838 control points, while the average distance of the people from the camera is 5.15 m. 

\textbf{Indoor sequence}
This sequence, counting 37 frames, simulates the presence of three to four people in a university classroom. We found about 679 control points in the scene, while the average distance from the camera is 4.85 m.

\subsection{Inter-Personal Distance Evaluation}

For each pair of people in each test frame, we first compute the inter-personal distance using the proposed strategy explained in Section \ref{inter-personal-distance}. Then, we measure the error between the latter and the one measured using the reference depth map obtained by the LiDAR sensor, showing how much our approach deviates from the metric measurements enabled by an accurate depth sensor. Table \ref{tab:distance_evaluation} illustrates the results of this evaluation, grouping outdoor and indoor environments, respectively. Specifically, to assess the quality of our approach, we report the results scaling the depth maps using three different types of control points: (a) depth inferred by a monocular SLAM system as outlined in our proposal, (b) the raw depth inferred by the LiDAR and (c) the raw depth inferred by the LiDAR constrained to a depth range similar to that of ARCore (not farther than about 10 meters).

Focusing on (a), we can notice how, in general, the deviation ranges from 0.27 cm (DPT-Hybrid), comparable with the accuracy of the ARCore control points, to a much higher value of more than 1 meter (MIDAS small). It is often related to the complexity of the depth network deployed, with more complex ones more accurate. Although this is not always true, on average, MIDAS and DPT perform overall better than others, with the lightweight PyD-Net model sometimes really close or better in the outdoor environment. By comparing the performance in the two scenarios, we can notice how only MIDAS can yield an accuracy in the inter-personal distance below half a meter considering any range between people. Limiting such an evaluation only for people at a distance below 3 meters, excluding one case (PyD-Net in Indoor), all the networks improve their performance by a significant margin with MIDAS and DPT models more prominently than others. This latter evaluation is particularly important for social distance monitoring since potential violations occur when people get closer. According to the evaluation reported in the table, MIDAS is the most effective network yielding an uncertainty of 33 cm on average.

Concerning results (b) and (c) in the table, it is interesting to analyze the framework behavior when providing more accurate control points. Indeed the ARCore depth accuracy is far from being perfect as yet observed in Section \ref{evaluation_sfm}. In configurations (b) and (c), although not consistently, most networks improve their performance. Sometimes the improvement is notable and higher when the sparse background data are closer, such as for PyD-Net yielding results almost overlapping with the DPT networks as reported in (C) for distances smaller than 3 m. With both LiDAR data distributions, the best accuracy considering distance below 3 m gets close to 20 cm, slightly better than deploying ARCore depth data. On the one hand, this fact highlights that the uncertainty of depth measurements needed to scale the network has a not marginal impact.  On the other hand, most of the depth uncertainty occurs due to the ill-posedness problem the monocular depth networks face, although some are undoubtedly more effective than others.

 \begin{table*}[t]
 \centering
 \scalebox{0.8}{
 \begin{tabular}{c}
 \begin{tabular}{l c ccc|ccc|ccc c ccc|ccc|ccc}
    & & \multicolumn{9}{c}{Outdoor} & & \multicolumn{9}{c}{Indoor} \\
    & & \multicolumn{3}{c|}{\cellcolor{green!30}Safe} &
    \multicolumn{3}{c|}{\cellcolor{yellow!30}Risky} &
    \multicolumn{3}{c}{\cellcolor{red!30}Dangerous} & & \multicolumn{3}{c|}{\cellcolor{green!30}Safe} &
    \multicolumn{3}{c|}{\cellcolor{yellow!30}Risky} &
    \multicolumn{3}{c}{\cellcolor{red!30}Dangerous} \\
    \cline{1-1} \cline{3-11} \cline{13-21} 
    Method & & P & R & F & P & R & F & P & R & F & & P & R & F & P & R & F & P & R & F \\ 
    \cline{1-1} \cline{3-11} \cline{13-21} 
    ours/PyD-Net & & 0.94 & 0.83 & 0.88 & 0.69 & 0.79 & 0.74 & 0.45 & 0.70 & 0.55 & & 0.53 & 0.81 & 0.64 & 0.47 & 0.36 & 0.41 & 0.83 & 0.28 & 0.42\\
    ours/MIDAS & & 0.91 & 0.87 & 0.89 & 0.73 & 0.70 & 0.71 & 0.47 & 0.75 & 0.58 & & 0.79 & 0.79 & 0.63 & 0.63 & 0.72 & 0.67 & 0.67 & 0.44 & 0.53 \\
    ours/MIDAS small & & 0.88 & 0.85 & 0.86 & 0.70 & 0.63 & 0.67 & 0.31 & 0.68 & 0.43 & & 0.59 & \underline{\textbf{0.89}} & \underline{\textbf{0.71}} & \underline{\textbf{0.68}} & 0.46 & 0.55 & \underline{\textbf{0.88}} & 0.47 & 0.61 \\
    ours/DPT-Hybrid & & 0.94 & \textbf{0.89} & 0.91 & \textbf{0.79} & 0.80 & 0.79 & \textbf{0.58} & 0.91 & \underline{\textbf{0.70}} & & \underline{\textbf{1.00}} & 0.43 & 0.60 & 0.55 & 0.61 & 0.58 & 0.46 & \underline{\textbf{1.00}} & 0.63 \\
    ours/DPT & & \underline{\textbf{0.95}} & \textbf{0.89} & \textbf{0.92} & \textbf{0.79} & \underline{\textbf{0.81}} & \textbf{0.80} & 0.51 & \underline{\textbf{0.92}} & 0.66 & & \underline{\textbf{1.00}} & 0.45 & 0.62 & 0.60 & \underline{\textbf{0.84}} & \underline{\textbf{0.70}} & 0.67 & 0.84 & \underline{\textbf{0.74}} \\
    \cline{1-1} \cline{3-11} \cline{13-21} 
    \cline{1-1} \cline{3-11} \cline{13-21} 
    (a) Homography-based & & 0.90 & 0.97 & \underline{\textbf{0.93}} & \underline{\textbf{0.89}} & 0.77 & \underline{\textbf{0.83}} & \underline{\textbf{0.65}} & 0.58 & 0.61 & & \xmark & \xmark & \xmark & \xmark & \xmark & \xmark & \xmark & \xmark & \xmark \\
    (b) Aghaei et al \cite{Aghaei_2021_WACV} & & 0.35 & \underline{\textbf{0.98}} & 0.51 & \xmark & \xmark & \xmark & \xmark & \xmark & \xmark & & \xmark & \xmark & \xmark & \xmark & \xmark & \xmark & \xmark & \xmark & \xmark \\
    \cline{1-1} \cline{3-11} \cline{13-21} 
    \cline{1-1} \cline{3-11} \cline{13-21} 
 \end{tabular}
 \\
 \\
 \end{tabular}
 }
 \caption{\textbf{Evaluation about risk detection.} Evaluation of pair-wise inter-personal distance estimation for predicting Safe, Risky and Dangerous situations w.r.t the reference depth provided by the LiDAR, averaged over all pairs and images. For each sequence, we highlight with \textbf{bold} the best result within our methods and with \underline{\textbf{underline and bold}} the best result overall. The tag \xmark{} means that the information is not available, either because the method cannot be applied at all due to the scene's constraints or it does not provide the desired metric.}
 \label{tab:violation_detection}
 \end{table*}

\subsection{Detecting violations} \label{detecting_violations}
In the previous experiment, we validated our framework by computing the average error, in meters, between the expected and predicted inter-personal distance among people. However, another critical question is: \textit{how many times the system fails in predicting violations of the necessary inter-personal distance?} This point is crucial because even if the depth prediction is partially wrong, social distancing could be effectively fulfilled. Moreover, the concept of risk depends on the context: distances lower than one meter could be crucial for some critical applications, but a too strict constraint in less critical ones. For this reason, we evaluate next the performance of the system with three different classes of risk, \colorbox{green!30}{Safe}, \colorbox{yellow!30}{Risky} and \colorbox{red!30}{Dangerous}, representing when the relative distance among two people is $> 2$ m, in the range $[1,2]$ m and $< 1$ m respectively.
We evaluate the Precision (P), the Recall (R) and the F1 score (F), in $[0,1]$ format, of all the depth models using these three classes of risk. Table \ref{tab:violation_detection} collects the outcome of this evaluation, on both outdoor and indoor data. We compare the various flavours of our approach to two ways to recover interpersonal distances by means of homography computation (a) finding the position of each person as the center of the side of the bounding box touching the floor (e.g. \cite{fabbri2020inter,gad2020vision_social_distance}) or (b) applying the method by Aghaei et al. \cite{Aghaei_2021_WACV}, automatically detecting inter-personal distances lower than 2m.

\begin{figure*}[t]
    \centering
    \renewcommand{\tabcolsep}{1pt}
    \begin{tabular}{cccccc}
        \includegraphics[width=0.16\textwidth]{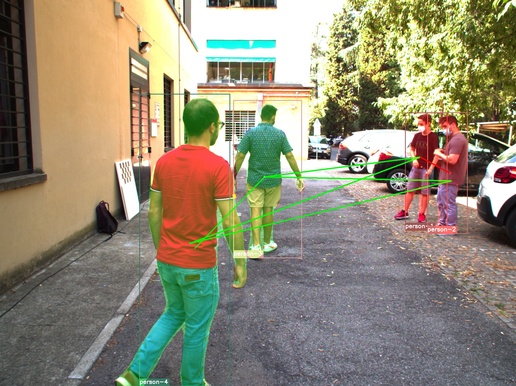} &
        \includegraphics[width=0.16\textwidth]{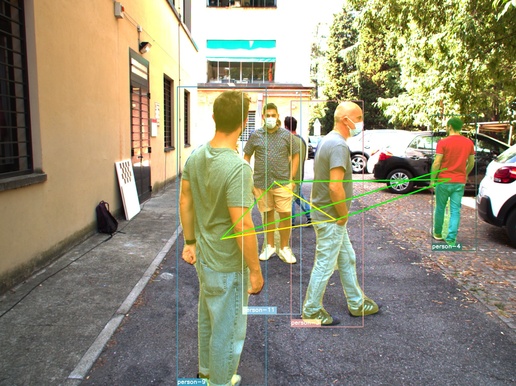} &
        \includegraphics[width=0.16\textwidth]{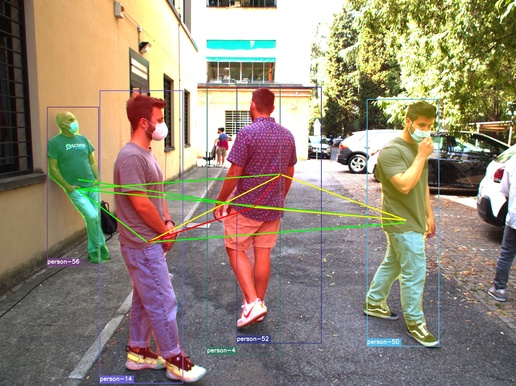} &
        \includegraphics[width=0.16\textwidth]{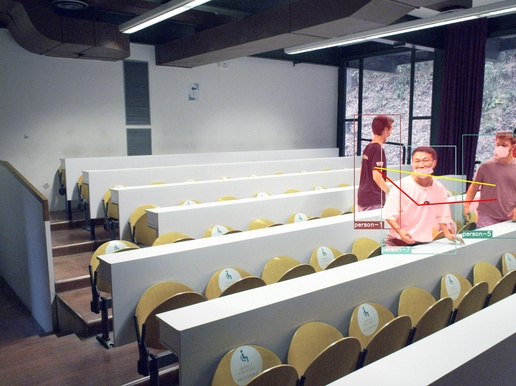} &
        \includegraphics[width=0.16\textwidth]{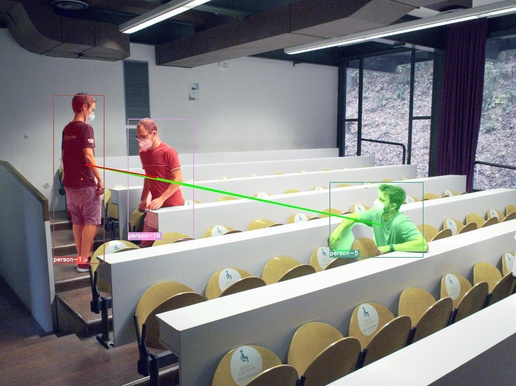} &
        \includegraphics[width=0.16\textwidth]{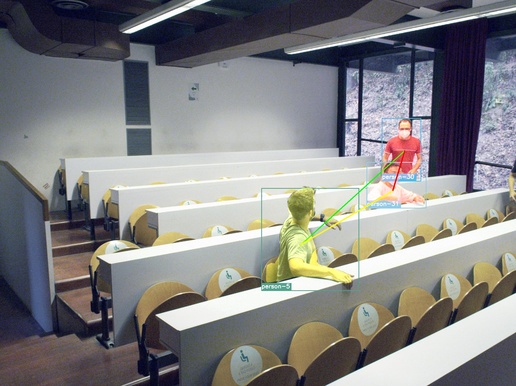} \\
        
    \end{tabular}
    \caption{\textbf{Qualitative results.} Examples from the collected scenes. The computed inter-personal distance is superimposed with a color overlay to the images -- green, yellow and red, mean \colorbox{green!30}{safe}, \colorbox{yellow!30}{risky} and \colorbox{red!30}{dangerous}.}
    \label{fig:qualitative}
\end{figure*}

Starting from outdoor samples (left), we notice how all models effectively distinguish safe situations, reaching a precision often superior to 0.90 and a recall between 0.83 and 0.89. DPT models allow for the best results, reaching in the worst case a 0.91 F1 score. MIDAS and PyD-Net perform similarly, reaching an F1 score slightly below 0.90. Excluding MIDAS small, the precision results are better than exploiting the homography (a) and vastly outperforms \cite{Aghaei_2021_WACV} (b), while always resulting worse in terms of recall. The F1 score achieved by our solution is almost equivalent, using DPT models, with (a), and much better than (b) \cite{Aghaei_2021_WACV}. The latter yields many false positives, often identifying violations of the social distancing constraint even when not occurring.

Analyzing risky situations, the precision for all networks drops at about 0.79 for DPT models and around 0.70 for the others. DPT models and MIDAS keep a similar score regarding the recall metric, while PyD-Net performs better and MIDAS small drops to 0.63. DPT variants achieve F1 around 0.80, still performing close to (a), achieving higher precision. The framework (b) by Aghaei et al. \cite{Aghaei_2021_WACV} is implemented to distinguish violations of the 2m distances, not allowing for finer processing and thus not able to detect risky and dangerous situations. 
Finally, all models' precision scores are generally significantly lower concerning dangerous situations, whereas recall metrics are higher, except PyD-Net than those observed for risky situations. (a) results are better in terms of precision, worse concerning the recall. The best F1 score overall is achieved by DPT variants.

Moving to the indoor scene, it is worth noticing how methods based on homography (a) and (b) cannot be computed here, as we can notice by looking at Fig. \ref{fig:dataset} since in this specific environment the ground plane is not visible. Nevertheless, our method can still reliably monitor social distancing since it does not require the ground plane to run.
We observe excellent precision scores for DPT models when dealing with safe cases, while other models are significantly less accurate. The opposite behaviour occurs for the recall. Surprisingly, the best F1 score  0.71 is achieved by MIDAS small, with MIDAS and PyD-Net following. Concerning risky conditions, in most cases, the performance degrades significantly, especially for lightweight models MIDAS small and PyD-Net. Finally, we can notice mixed results concerning dangerous situations with DPT models achieving the best results for recall and F1 metric, but outperformed by MIDAS small in precision. Looking at the recall, we can notice PyD-Net struggling, as for the risky case. 

In summary, the evaluation with the three classes of risk highlights that the models are pretty effective in handling safe or risky situations between people, especially in the former case. Nonetheless, they become more unreliable when facing dangerous situations with distances falling under one meter. DPT models achieve the best performance in most cases while lightweight models, PyD-Net and MIDAS small, further emphasize the outlined behavior. In general, our approach is competitive compared to homography computation (a) and much better than (b) \cite{Aghaei_2021_WACV} overcoming their main limitation -- the need for the ground plane to be visible and detectable.
We report some qualitative results concerning the outcome of the proposed framework in Fig. \ref{fig:qualitative}.

\subsection{Runtime analysis}

 Table \ref{tab:runtime_analysis} reports the performance of each component of the proposed overall framework (as average time per frame) on CPU and GPU. Moreover, we also report the same metric for the two competitors on the same hardware when feasible. The depth estimation models vary from more than 4 (DPT) to more than 60 FPS (PyD-Net and MIDAS small) on a quite outdated Nvidia GTX Titan X GPU. The YolactEdge instance segmentation network accounts for 0.055s on the same hardware. Focusing on the overall system, when using the same GPU, our approaches run at frame rate ranging from 9.4 FPS, with Pyd-Net and MIDAS small, to more than 3, with DPT, comparable to competitors. 
 With a pure CPU system, the runtime of the instance segmentation network becomes more prominent, especially compared to the fastest depth estimation networks. Nonetheless, the largest depth estimation models, such as DPT, account for even higher execution time. Our framework runs at a frame rate ranging from more than 2, with Midas small and Pyd-Net -- slightly slower than the homography-based approach, to 0.4, with DPT.  
Finally, we conclude by reporting that with an NVidia Jetson Nano, our framework runs at about 0.5 FPS using the fastest configurations, with power consumption below 10W, suggesting the practical deployment of our approach even on low-power embedded systems is around the corner.

\begin{table}[t]
    \centering
    \begin{tabular}{|c|c|c|c|}
        \hline
        \textbf{Method} & \textbf{Component} & \textbf{CPU} & \textbf{GPU}  \\
        \hline
        & PyD-Net & 0.033s & 0.016s\\
        \cline{2-4}
        & MIDAS small & 0.067s & 0.016s \\
        \cline{2-4}
        ours/ & DPT & 2.125s & 0.225s\\
        \cline{2-4}
        & YolactEdge & 0.349s & 0.055s\\
        & Remaining computations & 0.050s & 0.035s\\
        \hline
        Homography-based & YolactEdge & 0.349s & 0.055s\\
         & Remaining computations & 0.050s & 0.035s\\
        \hline
        Aghaei et al. \cite{Aghaei_2021_WACV} & Social-Distancing & -\footnotemark & 0.110s\\
        \hline
    \end{tabular}
    \vspace*{+2mm}
    \caption{\textbf{Runtime evaluation.} With images at resolution 826$\times$618, we measure the performance of each subsystem of our proposal, with different depth models and its competitors. The CPU and GPU employed are an AMD Ryzen 5900X and an Nvidia GTX Titan X. 
    }
    \label{tab:runtime_analysis}
\end{table}

\footnotetext{We could not install the CPU version of OpenPose on our hardware. Nevertheless, the authors claim that the default CPU version takes $\sim$0.2 images per second on Ubuntu ($\sim$50x slower than GPU) while the MKL version provides a roughly 2x speedup at $\sim$0.4 images per second.}

\section{Limitations}
The proposed system proved to be effective in monitoring inter-personal distances in non-critical applications. Although versatile and effective even when the ground plane of the scene is not visible or there are multiple planes, like when sensing a staircase, it might fail under certain conditions. In the following, we summarize the most notable ones.

\textbf{Monocular generalization.}
The monocular network itself represents a critical aspect. Even if we adopt networks with strong generalization capabilities, they might fail when misleading images, such as those acquired by unusual perspectives. In this case, the network could fail to predict the correct depth for the scene, and the scaling alone would not recover its correct geometry. However, as discussed in Section \ref{inter-personal-distance}, since we compute the centroids in the 2D space, partial failures in monocular predictions are tolerable for inter-personal distance computation as long as the depth of the centroid is correct. Even if the monocular network assigns a wrong depth to the head of the person (monocular depth image in Figure \ref{fig:monocular_failure}), the centroid is found on the body, where the depth is consistent. The problem outlined is well-known in literature, and in future works, it could be tackled employing learned refinement strategies \cite{yin2021learning} or performing in-domain fine-tuning.

The choice of using the centroid as defined in Section \ref{inter-personal-distance} may not be the best when searching for the social distance, although most existing approaches follow similar strategies \cite{fabbri2020inter,Aghaei_2021_WACV}. A stricter approach could be to use the nearest points between each pair of people. However, this method is computationally expensive and less robust to noise in depth computation. Future work could concern finding a proper way to retain robust measures even when deploying more complex approaches for people relative localization.

\textbf{Control points sourcing.}
Another issue concerns the control points needed to scale the predicted maps. A subset of at least two points must be visible at inference time. Thus, an extremely crowded environment might potentially result in a scene without visible control points. However, this seems very unlikely, and we never faced such situations in our experiments.
Additionally, control points could be detected on objects that might move, \eg{} cars in Figures \ref{fig:dataset} and \ref{fig:qualitative}. However, these points could be easily discarded at test time if properly detected, for instance, employing background subtraction techniques.

\textbf{People detection.}
Lastly, another source of errors is the people segmentation network. When the network fails in segmenting different people or, in the worst case, it misses a person; our framework would predict wrong inter-personal distances. Nonetheless, we would like to point out that this problem is shared with other vision-based social preserving applications and that it could be partially alleviated by exploiting temporal information from the video stream (\eg{} using the Kalman filter to forecast the next position of the centroid).

\section{Conclusions}
In this paper, we proposed a framework aimed at social distance monitoring from a single image. The system requires a single, static RGB camera, making it suitable for already deployed infrastructures such as surveillance cameras. In contrast to most other monocular approaches proposed in the literature, our framework is robust even when the ground plane is not visible, or there are multiple ones, such as in the presence of staircases. To retrieve the metric scale factor out of the output of monocular depth networks, we propose a simple yet effective strategy to source sparse depth points requiring just a smartphone with a single camera. Experimental results highlight that our proposal is effective, particularly as a versatile and cheap solution for non-critical applications. To the best of our knowledge, this is one of the first attempts to employ monocular depth networks in practical applications requiring metric distances.

\section*{Acknowledgments}
\scriptsize{This work was partially supported by Cloudif.ai within the project 'Intelligenza Artificiale per la realizzazione di un sistema informatico per la sicurezza della salute delle persone in contrasto a Covid-19 mediante Computer Vision e Machine Learning' funded by Regione Emilia-Romagna through program POR-FESR 2014-2020 concerning research and innovation aimed at facing the Covid-19 pandemic -- project code E81B20000600007.

We acknowledge with thanks Simone Amorati, Alessandro Musarella, Fabio Scagliarini, Amedeo Ravagli, and  Lorenzo Righi for preliminary activities concerning this work.}

\ifCLASSOPTIONcaptionsoff
  \newpage
\fi

\bibliographystyle{IEEEtran}
\bibliography{egbib}

\end{document}